\documentclass[10pt,letterpaper]{article}
\usepackage[top=0.85in,left=2.75in,footskip=0.75in]{geometry}

\usepackage{amsmath,amssymb}

\usepackage{changepage}

\usepackage{textcomp,marvosym}

\usepackage{cite}

\usepackage{nameref,hyperref}

\usepackage[right]{lineno}

\usepackage[nopatch=eqnum]{microtype}
\DisableLigatures[f]{encoding = *, family = * }

\usepackage[table]{xcolor}
\usepackage{multirow}
\usepackage{makecell}
\usepackage{booktabs}
\usepackage{nicematrix}
\usepackage{textgreek}

\usepackage{array}

\newcolumntype{+}{!{\vrule width 2pt}}

\newlength\savedwidth

\raggedright
\usepackage[aboveskip=1pt,labelfont=bf,labelsep=period,justification=raggedright,singlelinecheck=off]{caption}

\makeatletter
\renewcommand{\@biblabel}[1]{\quad#1.}
\makeatother

\usepackage{lastpage,fancyhdr,graphicx}
\usepackage{epstopdf}
\fancyheadoffset[L]{2.25in}
\fancyfootoffset[L]{2.25in}
\begin{document}
\vspace*{0.2in}

\begin{flushleft}
{\Large
\textbf\newline{LMFLOSS: A hybrid loss for imbalanced medical image classification} % Please use "sentence case" for title and headings (capitalize only the first word in a title (or heading), the first word in a subtitle (or subheading), and any proper nouns).
}
\newline
\\
Abu Adnan Sadi\textsuperscript{1*},
Labib Chowdhury\textsuperscript{2},
Nusrat Jahan\textsuperscript{1},
Mohammad Newaz Sharif Rafi\textsuperscript{1},
Radeya Chowdhury\textsuperscript{3},
Faisal Ahamed Khan\textsuperscript{2},
Nabeel Mohammed\textsuperscript{1}
\\
\bigskip
\textbf{1} Department of Electrical and Computer Engineering, North South University, Dhaka, Bangladesh
\\
\textbf{2} Giga Tech Limited, Dhaka, Bangladesh
\\
\textbf{3} City Hospital, Dhaka, Bangladesh
\bigskip

% Insert additional author notes using the symbols described below. Insert symbol callouts after author names as necessary.
% 
% Remove or comment out the author notes below if they aren't used.
%
% Primary Equal Contribution Note
%\Yinyang These authors contributed equally to this work.

% Additional Equal Contribution Note
% Also use this double-dagger symbol for special authorship notes, such as senior authorship.
%\ddag These authors also contributed equally to this work.

%\textcurrency Current Address: Dept/Program/Center, Institution Name, City, State, Country % change symbol to "\textcurrency a" if more than one current address note
% \textcurrency b Insert second current address 
% \textcurrency c Insert third current address


% Group/Consortium Author Note
%\textpilcrow Membership list can be found in the Acknowledgments section.

% Use the asterisk to denote corresponding authorship and provide email address in note below.
* abu.sadi05@northsouth.edu, labib.chowdhury@gigatechltd.com, mohammad.newaz@northsouth.edu, nusrat.jahan13@northsouth.edu,
chowdhuryd944@gmail.com, faisal.cse06@gigatechltd.com, nabeel.mohammed@northsouth.edu

\end{flushleft}
\section*{Abstract}
With advances in digital technology, the classification of medical images has become a crucial step for image-based clinical decision support systems. Automatic medical image classification represents a pivotal domain where the use of AI holds the potential to create a significant social impact. However, several challenges act as obstacles to the development of practical and effective solutions. One of these challenges is the prevalent class imbalance problem in most medical imaging datasets. As a result, existing AI techniques, particularly deep-learning-based methodologies, often underperform in such scenarios. In this study, we propose a novel framework called Large Margin aware Focal (LMF) loss to mitigate the class imbalance problem in medical imaging. The LMF loss represents a linear combination of two loss functions optimized by two hyperparameters. This framework harnesses the distinct characteristics of both loss functions by enforcing wider margins for minority classes while simultaneously emphasizing challenging samples found in the datasets. We perform rigorous experiments on three neural network architectures and with four medical imaging datasets. We provide empirical evidence that our proposed framework consistently outperforms other baseline methods, showing an improvement of 2\%-9\% in macro-f1 scores. Through class-wise analysis of f1 scores, we also demonstrate how the proposed framework can significantly improve performance for minority classes. The results of our experiments show that our proposed framework can perform consistently well across different architectures and datasets. Overall, our study demonstrates a simple and effective approach to addressing the class imbalance problem in medical imaging datasets. We hope our work will inspire new research toward a more generalized approach to medical image classification. Our source code is publicly available at \url{https://github.com/Adnan-Sadi/LMFLOSS}.

\nolinenumbers

% Use "Eq" instead of "Equation" for equation citations.
\section*{Introduction}
The recent developments of AI, specifically in neural network-based computer vision techniques, have enabled the possibility of creating automatic intelligent diagnostic tools based on medical images to achieve human-level performance \cite{Shen2017}. Medical image analysis has shown considerable potential when using supervised learning, where complex neural network models are trained on large volumes of labeled data \cite{Marleen2016}. However, such systems are mostly trained on images of frequently occurring diseases, which limits their effectiveness. It is common for medical imaging datasets to contain a significantly lower number of samples of rare diseases than samples of common ones. This class imbalance causes the neural network models to become biased and perform poorly on the minority classes, barring their use as assistive technologies to human specialists \cite{Fotouhi2019}. This study aims to address this class imbalance challenge.

We can categorize previous studies to address the class imbalance issue into two broad approaches: data-centric strategies and algorithmic strategies. Data-centric strategies encompass different data sampling approaches used to tackle data imbalance. Oversampling and undersampling are the two most popular data-centric approaches. The random undersampling method balances the data by eliminating samples from the majority classes \cite{Seiffert2010}. In contrast, oversampling adds artificially generated or duplicated data to the minority classes \cite{Tran2022}. Another well-known oversampling technique called SMOTE (synthetic minority over-sampling technique) \cite{Elreedy2019} creates "synthetic" samples for minority class rather than simply replicating them. Due to the simplicity and popularity of such sampling methods, several studies in the medical domain have taken similar approaches to address the imbalance problem of medical datasets\cite{Rahman2013, Qu2020, Dubey2014, Sekuboyina2017}.

Even though data-centric sampling techniques can be effective in certain scenarios, they are not always feasible. For instance, undersampling could potentially lead to the removal of valuable data \cite{kotsiantis2005}, whereas oversampling could increase the number of duplicate samples in the training data, which could lead to overfitting \cite{Drummond2003}. Moreover, oversampling increases the number of training data, leading to longer training periods. Similarly, the SMOTE oversampling technique also has its own set of challenges including the risk of worsening the overfitting issue by oversampling noisy data or oversampling less informative samples \cite{Barua2014}. In a more recent study,  Misuk Kim and Kyu-Baek Hwang \cite{Kim2022} performed a comprehensive analysis of seven sampling methods to assess the effectiveness of sampling methods for classifying imbalanced data. They observed that the application of sampling was more likely to deteriorate the performance of a classifier rather than improve it.

In contrast to data-centric strategies, there exist various solutions that primarily focus on algorithm-centric methods to address the issue of class imbalance. Cost-sensitive learning is one such method that has been widely utilized in the medical domain \cite{Mienye2021, SUN20073358, Mostefa2020}. Cost-sensitive learning is applied by introducing class weights to the loss functions. Higher weights are given to the minority classes so that the loss functions can direct the models to concentrate more on accurately identifying the minority classes. On the other hand, some studies in the medical domain have also utilized novel network architectures \cite{Bria2020, Sakamoto2018, Chen2022, Fatima2022} to mitigate the class imbalance problem. Additionally, several researchers also proposed novel loss functions specifically designed to address the class imbalance problem, such as Focal loss \cite{Lin2017}, Label-Distribution-Aware Margin(LDAM) loss \cite{Cao2019}, and Class-Balanced loss \cite{Cui2019}. Multiple research works in the medical domain have applied loss function-based methods to address the imbalance issue in medical datasets \cite{Lei2020, rajaraman2021novel, bennett2022imbalance, tran2019improving, Roy2022}.

\begin{figure*}[!h]
\includegraphics[width=\linewidth]{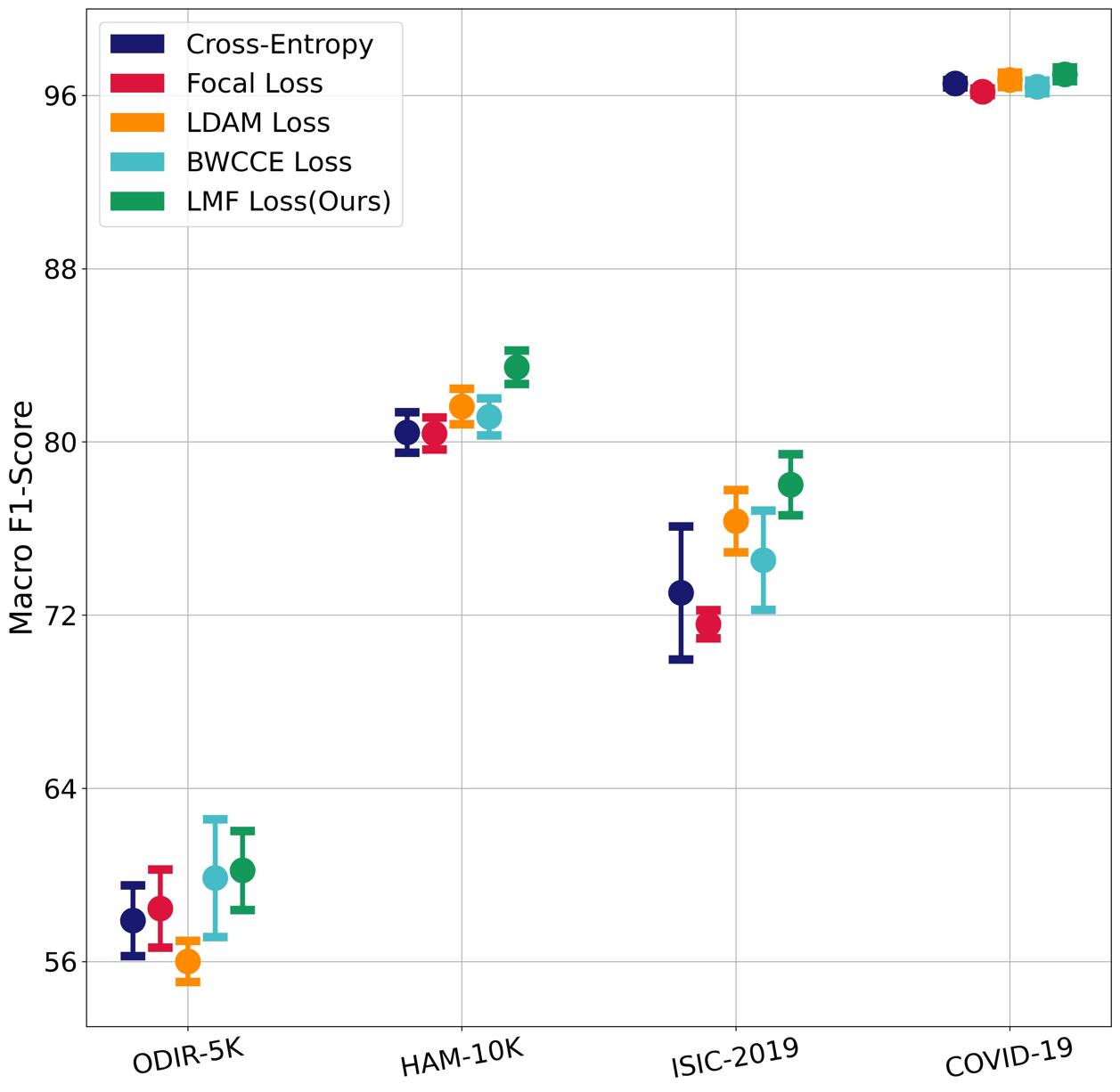}
\caption{{\bf Macro-f1 scores of the proposed method and compared to four other pre-existing techniques for four different medical image datasets.}
Each error bar depicts the mean of the macro-f1 scores obtained from three different network architectures, along with its average deviation. The proposed LMF-loss achieves higher average macro-f1 scores for all four datasets.}
\label{fig1}
\end{figure*}

In our study, we explore several loss function-based approaches for addressing the class imbalance issue in the medical imaging domain. In addition, we propose a simple yet effective loss framework that can utilized to mitigate the class imbalance problem in medical datasets. The key contributions of this paper are summarized as follows:

\begin{enumerate}
	\item{We propose a novel framework called Large Margin aware Focal(LMF) loss, which combines two different loss functions in a hybrid framework and jointly optimizes them. This loss framework dynamically takes hard samples into consideration and, depending on the class distribution, simultaneously imposes larger margins on the minority classes from the decision boundary.}
 
	\item{We demonstrate the effectiveness of our proposed framework by providing a thorough performance comparison with four other existing loss functions. In addition, all the experiments were conducted on three different neural network architectures.}
 
	\item{We demonstrate the robustness of our proposed method by conducting extensive experiments on four popular datasets from three different medical imaging domains. The selected datasets include: Ocular Disease Intelligent Recognition(ODIR-5K) \cite{odir2019}, the Human Against Machine(HAM) \cite{Tschandl2018}, the International Skin Imaging Collaboration(ISIC)-2019 \cite{Gutman2016}, and the COVID-19 Radiography Dataset \cite{Chowdhury2020, Rahman2021}. The ODIR-5K and COVID-19 Radiography datasets contain images from the color fundus photography and chest X-ray domains, respectively. On the other hand, the HAM-10K and ISIC-2019 datasets contain skin images for skin lesion identification. In contrast, most of the prior studies related to medical data imbalance primarily focus on a single dataset or multiple datasets from the same medical imaging domain.}
    
    \item{On all four datasets, the proposed method achieved significant performance improvement in the macro-f1 score when compared to other baselines \textbf{(see Fig \ref{fig1})}. We provide detailed quantitative results of the performance comparison with multiple evaluation metrics. We also provide a comparative analysis class-wise of f1 scores. Finally, we provide qualitative results by performing Grad-CAM \cite{Selvaraju2016} attention map analysis.}
\end{enumerate}

% Place figure captions after the first paragraph in which they are cited.
% For figure citations, please use "Fig" instead of "Figure".

\section*{Materials and methods}

\subsection*{Baselines}
In this section, we briefly discuss the four loss functions we used as baselines for our study.

\subsubsection*{Categorical cross-entropy loss}
Cross-entropy is used for measuring the difference between two probability distributions for a particular set of instances. The Categorical Cross-Entropy (CCE) loss is widely used for multi-class classification problems. It measures the difference between the predicted probabilities and the ground truth labels. The standard categorical cross-entropy loss can be represented as follows:

\begin{equation}
L_{CCE} = -\frac{1}{N} * \sum_{i=1}^{N} \sum_{j=1}^{K} y_{i,j} * log(p_{i,j})
\label{eq:cce}
\end{equation}

Here, y\textsubscript{i,j} is the ground truth label for the i-th sample of class j, p\textsubscript{i,j} is the probability that the i-th sample belongs to class j, K is the total number of classes, and N is the total number of training samples.

However, the CCE loss does not account for the class imbalance issue during loss calculation. As a result, researchers have proposed adding a weight variable to the standard CCE formula \cite{Ho2020}. This approach is also known as `cost-sensitive learning', which adds class weights to the conventional loss functions. The weighted version of the Categorical Cross-Entropy (WCCE) can be expressed as follows:

\begin{equation}
L_{WCCE} = -\frac{1}{N} * \sum_{i=1}^{N} \sum_{j=1}^{K} w_j * y_{i,j} * log(p_{i,j})
\label{eq:wcce}
\end{equation}

Here, w\textsubscript{j} is the assigned weight for class j. Class weights are generally defined as the inverse ratio of the number of images present in each class. Higher weights get assigned to minority classes, thus increasing the loss value when models misclassify a sample from the minority class. 

\subsubsection*{Balanced weighted categorical cross-entropy loss}
In extreme circumstances of imbalance, the minority classes may have much fewer images than the other classes. As a result, the minority classes may have very high weight values, consequently making the model more biased toward the minority classes. This disparity in weight values can result in a fluctuation in the model's performance for other classes. To mitigate this issue, the authors of \cite{Roy2022} proposed the Balanced Weighted Categorical Cross Entropy (BWCCE) loss.

The authors followed the same intuition of the class weights being inversely proportional to the distribution of images in each class. However, they defined the weights using the concept of probability; which ensured the sum of the weight values assigned for all the classes would always equal 1. They also showed that using this method, the weight value of the minority classes would not deviate too much from the other classes, even if the imbalance is extreme. The formula for BWCCE loss is the same as \textbf{Eq (\ref{eq:wcce})}. But the authors define the weight w\textsubscript{j} with the following formula:

\begin{equation}
w_j = \frac{1}{K-1} (1 - \frac{n_j}{\Sigma_{j} n_j})
\label{eq:bwcce}
\end{equation}

Here, K is the total number of classes and (K\textgreater1). n\textsubscript{j} is the total number of samples in class j, and \textSigma\textsubscript{j}n\textsubscript{j} is the total number of samples in the dataset.

\subsubsection*{Label distribution aware margin loss}
Another work to mitigate the class imbalance issue was proposed by authors from \cite{Cao2019} called Label-Distribution Aware Margin(LDAM) loss. They suggested regularizing the minority classes more strongly than the majority classes to decrease their generalization error. This way, the loss function maintains the model's capacity to learn the majority classes and emphasize the minority classes. The LDAM loss focuses on the minimum margin per class and obtaining per-class and uniform label test error instead of encouraging the large margins of the majority classes' training samples from the decision boundary. In other words, it encourages comparatively larger margins for the minority classes. The authors from \cite{Cao2019} proposed the formula for getting a class-dependant margin for multiple classes 1,...,k as:

\begin{equation}
\gamma_j= \frac{C}{n_j^{1/4}}\label{eq:ldam_margin}
\end{equation}

Here j \text{\(\in\)} \{1,...,k\} is a particular class, n\textsubscript{j} is the number of samples in that class, and C is a constant. Now, let's consider x as a particular example and y as the corresponding label for x. Let an example be (x, y) and a model f. Considering $z_y = f(x)_y$ denotes the model's output for that particular sample. Let \( u = e^{z_y - \Delta_y}\), where \(\Delta_j = \frac{C}{n_j^{1/4}}\),  for \( j \: \in \: \{1,...,k\}\). So, the defined LDAM loss is given in \textbf{Eq (\ref{eq:ldam})}:

\begin{equation}
L_{LDAM}((x,y),f) = -log\frac{u}  
{u + \sum_{j\neq y}e^{z_y}}\label{eq:ldam}
\end{equation}

\subsubsection*{Focal loss}
The main drawback of using Cross-entropy loss in an imbalanced classification problem is that it insists on equal learning across all the classes. Such learning has a negative impact on classification performance as the class distributions are highly imbalanced. Focal loss \cite{Lin2017} mitigates this issue by down-weighting the samples that are easy for the model to identify. Authors of focal loss modified the cross-entropy loss function to focus more on samples that are hard to classify. This is achieved by down-weighting the easy samples and up-weighting the hard samples present in the dataset. As a result, the model focuses more on the hard samples, which are usually from the minority classes. The focal loss in a multi-class classification setting is defined as \textbf{Eq (\ref{eq:focalloss})}:

\begin{equation}\label{eq:focalloss}
FL(p_t)= -(1-p_t)^\gamma log(p_t)
\end{equation}

Here, p\textsubscript{t} is the predicted probability score of the model, and \textgamma~ is the focusing parameter that can be tuned. A higher value of the \textgamma~ lowers the loss of the easy samples, which enables the model to turn its attention toward hard samples. When \textgamma~ = 0, the loss function becomes the standard cross-entropy loss. For our study, we used \textgamma~ = 1.5, which produced relatively better results on the selected datasets.

The authors also proposed an \textalpha-balanced variant of the focal loss, which introduced the weighting factor \textalpha~ to the loss function. For our study, we used the margin values obtained from \textbf{Eq (\ref{eq:ldam_margin})} as the weighting factor \textalpha. The \textalpha-balanced focal loss is defined as:

\begin{equation}
FL(p_t)= -\alpha_t (1-p_t)^\gamma log(p_t)\label{eq:focalloss_2}
\end{equation}

\subsection*{Large margin aware focal loss}
Focal loss creates a mechanism to give more emphasis to samples that are difficult for the model to classify by down-weighting the easy samples, consequently shifting the model's focus towards difficult samples. Quite often, samples from the minority classes would fall into this category. On the other hand, the LDAM loss calculates the margin by considering the class distribution of the dataset. It assigns a larger margin to the minority class from the decision boundary, which helps the model to focus more on the minority classes. Unlike the focal loss, the LDAM loss does not consider individual samples.

We hypothesized that simultaneously leveraging the two most unique features of the focal and LDAM loss could yield effective results compared to using each one individually. Our proposed Large Margin aware Focal (LMF) loss is thus a linear combination of Focal loss and LDAM weighted by two hyperparameters. As a result, the proposed hybrid framework can impose greater margins from the decision boundary based on the class distribution and can also take into account the harder samples that are present in the datasets. We add two hyperparameters to the proposed framework in order to adjust and control the influence of the two loss functions present within the LMF loss.

Using \textbf{Eq (\ref{eq:ldam})} and \textbf{Eq (\ref{eq:focalloss_2})}, the LMF loss is expressed by the following formula:
\begin{equation}
L_{LMF} = \alpha (-log\frac{u} {u + \sum_{j\neq y}e^{z_y}}) 
+ \beta (-\alpha_t (1-p_t)^\gamma log(p_t)) \label{eq:lmf}
\end{equation}

Here, \textalpha~ and \textbeta~ are constants and considered hyperparameters that can be adjusted. Thus, our proposed method jointly optimized two separate loss functions in a single framework. The outputs from the last fully connected layer of the model were used to calculate both the LDAM and Focal loss values. The hyperparameters were then adjusted to balance the influence of each loss function. From trial and error, we found that setting the \textalpha~ and \textbeta~ values between 0.5 to 2.0 yielded the best results. We present a detailed analysis of the hyperparameter values later in the results section.

\subsection*{Datasets}

\begin{figure}[!h]
\includegraphics[width=\linewidth]{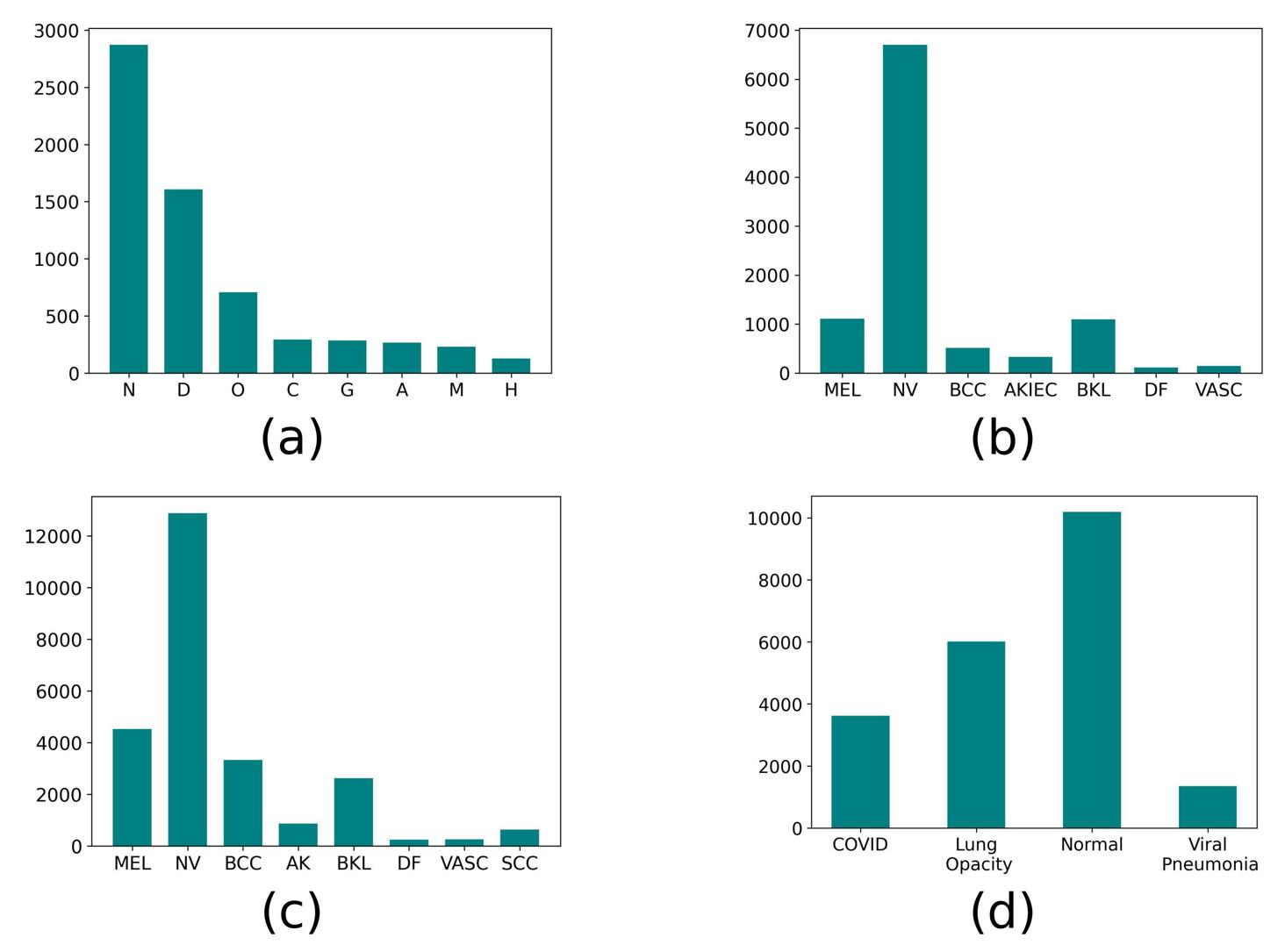}
\caption{{\bf Per-class image distribution of all four datasets.}
(a) ODIR-5K, (b) HAM-10K, (c) ISIC-2019, and (d) COVID-19 Radiography.}
\label{fig2}
\end{figure}

In this study, we performed our experiments on four different medical image datasets; the Ocular Disease Intelligent Recognition(ODIR) dataset \cite{odir2019}, the Human Against Machine (HAM) dataset \cite{Tschandl2018}, the International Skin Imaging Collaboration(ISIC-2019) dataset \cite{Gutman2016}, and COVID-19 Radiography Dataset \cite{Chowdhury2020, Rahman2021}. All four datasets were highly imbalanced (\textbf{see Fig \ref{fig2}}), making them perfectly suitable for our study.

We divided all four datasets into three sets; training set, validation set, and test set. For this study, the train, validation, and test set split ratio was set to 70:15:15. While splitting the datasets, we ensured that the per-class image distribution for each set was the same as the per-class image distribution of the whole dataset. Detailed information about the training, validation, and test sets for each dataset is given in \textbf{Table \ref{table:tab1}}. Prior to training, we resized the dimensions of all images to 224x224.

\begin{table}[!ht]
\caption{\textbf{Number of samples in Training, Validation, and Test Sets.}}
\centering
\begin{tabular}{|l|l|l|l|}

\hline
\textbf{Dataset} & \textbf{Training} & \textbf{Validation} & \textbf{Test}\\
\hline
ODIR & 4,474 & 959 & 959\\ \hline

HAM & 7,011 & 1,502 & 1,502\\ \hline

ISIC & 17,733 & 3,799 & 3,799\\ \hline

Covid-19 & 14,815 & 3,175 & 3,175\\
\hline

\end{tabular}
\label{table:tab1}
\end{table}

\subsection*{Models}
In this study, we used three different pre-trained neural network architectures, ResNet50 \cite{He2015}, EfficientNetV2 \cite{Tan2021}, and DenseNet121 \cite{Huang2016}. The models were pre-trained on the Imagenet \cite{Deng2009} 1000 class dataset. All three of these models are available on the PyTorch library.

\subsection*{Training parameters}
We chose a specific set of training parameters to do a comparative analysis of the performance of the four existing loss functions and our proposed framework. We opted to train each model for 100 epochs with a batch size of 32. We selected the Adam optimizer with a learning rate of 0.0001 and a scheduler that decayed the learning rate by a factor of 0.1 every 30 epochs. Also, we added a weight decay of 0.0005 to the optimizer to implement L2 regularization. We applied these hyperparameters and optimization settings in all of the experiments performed to compare the performance of the five different methods used in this study.

\subsection*{Evaluation metrics}
Along with the accuracy, precision, and recall, we have also used the macro f1 score to evaluate the performance of the models. The f1 score is the harmonic mean of precision and recall, whereas the macro f1 score is the arithmetic mean of per-class f1 scores for a more appropriate measurement of model performance on class-imbalanced data \cite{weiss2013}.

\begin{enumerate}
\item \textbf{Accuracy:} Accuracy is the most common method for evaluating classification models. It calculates the number of accurate predictions compared to the total number of labels. Accuracy is a very well-known method for model evaluation. But, it is not a suitable metric for imbalanced datasets.

\begin{equation}
Accuracy=\frac{TP + TN}{TP + TN + FP + FN}\label{eq_accuracy}
\end{equation}

Here, TP (True Positive) refers to a set of positive characteristics appropriately identified as such, while TN (True Negative) refers to a set of negative characteristics correctly identified as such. On the other hand, FP (False Positive) refers to characteristics that are actually negative but are projected to be positive, and FN (False Negative) refers to characteristics that are actually positive but are predicted to be negative.

\item \textbf{Precision:} Precision measures the proportion of the correct positive predictions compared to all the positive predictions that the model made. Precision for a label is defined as the number of true positives divided by the number of predicted positives. It is a suitable evaluation metric when we want to reduce the number of False Positives.

\begin{equation}
Precision=\frac{TP}{TP + FP}\label{eq_precision}
\end{equation}

\item \textbf{Recall:} Recall measures the proportion of actual positives that were predicted correctly by the model. Recall for a label is defined as the number of true positives divided by the total number of actual positives. It is a suitable evaluation metric when we want to reduce the number of False Negatives.

\begin{equation}
Recall=\frac{TP}{TP + FN}\label{eq_recall}
\end{equation}

\item \textbf{Macro F1:} F1 Score is a measure that combines Precision and Recall metrics. When both FP and FN are equally important, the f1 measure is a good choice.

\begin{equation}
F1 =2\times\frac{Precision\times Recall}{Precision+Recall}\label{eq_f1}
\end{equation}

For our study, we used the macro-f1 score, which is simply calculated by averaging the per-class f1 scores obtained from the model. Considering the total number of classes as n, the formula for the macro-f1 score can be written as:

\begin{equation}
Macro\;F1 = \frac{\sum_{i=1}^{n} F1_i}{n} 
\label{eq_macrof1}
\end{equation}

\end{enumerate}

\subsection*{Tools and libraries}
We used the PyTorch machine learning framework to train and test the models used in this study. Additionally, we used Python libraries such as Seaborn, Matplotlib, OpenCV, and Scikit-learn to generate the visualizations of the results. In particular, we used the M3d-CAM \cite{Gotkowski2021} PyTorch Library to generate the attention maps from the trained models.

\section*{Results and discussion}
\subsection*{Performance comparison of proposed LMF loss with baselines} \label{sec:lmfloss_results}
In this study, we conducted a comprehensive evaluation of four baseline loss functions and our proposed framework with three different convolutional neural network (CNN) architectures. Specifically, we trained the four selected datasets using five methods: Cross-Entropy loss, Focal loss, LDAM loss, BWCCE loss, and our proposed LMF loss. Our goal was to assess the effectiveness of the methods across diverse datasets and network architectures.  \textbf{Tables \ref{tab:table2}, \ref{tab:table3}, and \ref{tab:table4}} illustrate the results obtained from all the experiments. 

\begin{table}[!ht]
\centering
\caption{
{\bf Performance comparison of the proposed LMF-loss to other baseline loss functions on EfficientNetV2.}}
\begin{NiceTabular}{l|l|c|c|c|c}
\toprule
\textbf{Dataset} & \textbf{Loss} & \textbf{Accuracy} & \textbf{Precision} & \textbf{Recall} & \textbf{Macro F1}\\ 
\midrule

ODIR-5K & 
CCE loss & 67.64 & 65.55 & 58.21 & 60.34\\ \cmidrule{2-6}
& Focal loss & 67.36 & 62.75 & 60.23 & 60.84\\ \cmidrule{2-6}
& LDAM loss & 66.01 & 58.78 & 57.54  & 57.43\\ \cmidrule{2-6}
& BWCCE loss & 68.51 & 66.40 & \textbf{61.98} & \textbf{63.41}\\ \cmidrule{2-6}
& LMF loss \textbf{(ours)} & \textbf{69.86}  & \textbf{67.51} & 61.23 & 62.94\\ 
\midrule

HAM-10K & 
CCE loss & 89.65 & 85.64 & 78.80 & 81.85\\ \cmidrule{2-6}
& Focal loss & 88.55 & 82.46 & 79.81 & 80.75\\ \cmidrule{2-6}
& LDAM loss  & 89.48 & 84.40 & 80.34  & 82.12\\ \cmidrule{2-6}
& BWCCE loss & 90.01 & 84.53 & 80.64 & 81.99\\ \cmidrule{2-6}
& LMF loss \textbf{(ours)} & \textbf{91.21} & \textbf{86.55}  & \textbf{83.27} & \textbf{84.61}\\
\midrule

ISIC-2019 & 
CCE loss & 83.91 & 77.66  & 73.68 & 75.46\\ \cmidrule{2-6}
& Focal loss  & 80.52 & 71.62 & 72.23 & 71.73\\ \cmidrule{2-6}
& LDAM loss  & 85.36 & 80.57& 76.85& 78.50\\ \cmidrule{2-6}
& BWCCE loss  & 83.68 & 78.96 & 76.41 & 77.37\\ \cmidrule{2-6}
& LMF loss \textbf{(ours)} & \textbf{86.21} & \textbf{82.04}  & \textbf{78.79} & \textbf{80.15}\\
\midrule

Covid-19 & 
CCE loss  & 95.87 & 97.06 & 96.59 & 96.80\\ \cmidrule{2-6}
& Focal loss  & 95.37 & 95.90 & 96.09 & 95.99\\ \cmidrule{2-6}
& LDAM loss  & 96.38 & 97.15 & 97.10 & 97.11\\ \cmidrule{2-6}
& BWCCE loss  & 95.97 & 97.00 & 96.68 & 96.84\\ \cmidrule{2-6}
& LMF loss \textbf{(ours)} & \textbf{96.66} & \textbf{97.42}  & \textbf{97.36} & \textbf{97.38}\\
\bottomrule
\end{NiceTabular}
\begin{flushleft} In almost all cases, the proposed LMF loss framework outperformed other baselines for all four evaluation metrics. One exception was the ODIR-5k dataset, where the BWCCE loss outperformed the LMF loss on the recall and macro f1 metrics.
\end{flushleft}
\label{tab:table2}
\end{table}

\begin{table}[!ht]
\centering
\caption{
{\bf Performance comparison of the proposed LMF-loss to other baseline loss functions on ResNet-50.}}
\begin{NiceTabular}{l|l|c|c|c|c}
\toprule
\textbf{Dataset} & \textbf{Loss} & \textbf{Accuracy} & \textbf{Precision} & \textbf{Recall} & \textbf{Macro F1}\\ 
\midrule

ODIR-5K 
& CCE loss & 63.71  & \textbf{63.73} & 52.05& 55.79\\ \cmidrule{2-6}
& Focal loss & 61.63 & 56.26 & \textbf{56.65} & 55.75\\ \cmidrule{2-6}
& LDAM loss & 64.13 & 57.15 & 55.50  & 55.16\\ \cmidrule{2-6}
& BWCCE loss & 62.77 & 59.16 & 54.70 & 55.78\\ \cmidrule{2-6}
& LMF loss \textbf{(ours)} & \textbf{64.86}  & 62.58 & 56.12 & \textbf{57.73}\\
\midrule

HAM-10K & 
CCE loss & 87.82 & 81.97 & 78.43 & 79.95\\ \cmidrule{2-6}
& Focal loss & 86.48 & 76.41 & \textbf{82.74} & 79.28\\ \cmidrule{2-6}
& LDAM loss  & 88.95 & 81.06   & 80.37 & 80.41\\ \cmidrule{2-6}
& BWCCE loss & 88.15 & 82.72 & 78.57 & 79.88\\ \cmidrule{2-6}
& LMF loss \textbf{(ours)}& \textbf{89.35} & \textbf{83.89}  & 81.19 & \textbf{82.43}\\
\midrule

ISIC-2019 & 
CCE loss & 80.68 & 72.94  & 65.72 & 68.42\\ \cmidrule{2-6}
& Focal loss  & 80.28 & 70.39 & 71.61 & 70.60\\ \cmidrule{2-6}
& LDAM loss   & 81.97 & 74.62 & \textbf{76.90} & 75.64\\ \cmidrule{2-6}
& BWCCE loss  & 81.42 & 75.20 & 68.35 & 71.11\\ \cmidrule{2-6}
& LMF loss \textbf{(ours)}& \textbf{84.15} & \textbf{79.53}  & 76.29 & \textbf{77.68}\\
\midrule

Covid-19 & 
CCE loss & 95.75 & \textbf{96.74} & 96.18 & 96.44\\ \cmidrule{2-6}
& Focal loss  & 95.31 & 96.06 & 96.23 & 96.13\\ \cmidrule{2-6}
& LDAM loss  & 95.65 & 96.23 & 96.24 & 96.22\\ \cmidrule{2-6}
& BWCCE loss  & 95.50 & 95.97 & 96.31 & 96.12\\ \cmidrule{2-6}
& LMF loss \textbf{(ours)} & \textbf{95.81} & 96.66 & \textbf{96.42} & \textbf{96.50}\\
\bottomrule
\end{NiceTabular}
\begin{flushleft} The proposed LMF-loss outperformed other baselines in most cases, particularly when considering the accuracy and macro f1 metrics. In some instances, the LMF loss obtained the second-best precision and recall scores.
\end{flushleft}
\label{tab:table3}
\end{table}

\begin{table}[!ht]
\centering
\caption{
{\bf Performance comparison of the proposed LMF-loss to other baseline loss functions on DenseNet-121.}}
\begin{NiceTabular}{l|l|c|c|c|c}
\toprule
\textbf{Dataset} & \textbf{Loss} & \textbf{Accuracy} & \textbf{Precision} & \textbf{Recall} & \textbf{Macro F1}\\ 
\midrule

ODIR-5K & 
CCE loss & 65.28 & 61.14 & 56.02 & 57.53\\ \cmidrule{2-6}
& Focal loss & 64.86 & 60.49 & 58.19 & 58.75\\ \cmidrule{2-6}
& LDAM loss & 62.98 & 57.71 & 55.80 & 55.44\\ \cmidrule{2-6}
& BWCCE loss & \textbf{67.36} & \textbf{66.71} & \textbf{59.22} & \textbf{60.39}\\ \cmidrule{2-6}
& LMF loss \textbf{(ours)} & 66.32 & 65.14 & 58.84 & 59.97\\
\midrule

HAM-10K & 
CCE loss & 88.95 & 79.32 & 79.85 & 79.51\\ \cmidrule{2-6}
& Focal loss  & 88.48 & 84.00 & 79.04 & 81.15\\ \cmidrule{2-6}
& LDAM loss   & 89.21 & 83.12 & 82.18 & 82.40\\ \cmidrule{2-6}
& BWCCE loss  & 89.28 & 81.47 & 82.38 & 81.61\\ \cmidrule{2-6}
& LMF loss \textbf{(ours)} & \textbf{89.75} & \textbf{84.62}  & \textbf{82.77} & \textbf{83.31}\\
\midrule

ISIC-2019 & 
CCE loss & 82.78 & 76.78 & \textbf{73.87} & 75.20\\ \cmidrule{2-6}
& Focal loss  & 81.86 & 76.52 & 69.53 & 72.40\\ \cmidrule{2-6}
& LDAM loss   & 83.18 & 77.04 & 73.16 & 74.89\\ \cmidrule{2-6}
& BWCCE loss  & 83.76 & 77.90 & 72.89 & 75.14\\ \cmidrule{2-6}
& LMF loss \textbf{(ours)}  & \textbf{84.29} & \textbf{81.72}  & 72.39 & \textbf{76.27}\\
\midrule

Covid-19 & 
CCE loss & 95.62 & 96.42 & 96.42 & 96.41\\ \cmidrule{2-6}
& Focal loss  & 95.34 & 96.35 & 96.45 & 96.40\\ \cmidrule{2-6}
& LDAM loss   & 95.97 & 96.81 & 96.84 & 96.82\\ \cmidrule{2-6}
& BWCCE loss  & 95.50 & 96.31 & 96.25 & 96.24\\ \cmidrule{2-6}
& LMF loss \textbf{(ours)}  & \textbf{96.35} & \textbf{97.06} & \textbf{97.07} & \textbf{97.06}\\

\bottomrule
\end{NiceTabular}
\begin{flushleft} In almost all cases, the proposed LMF-loss outperformed other baselines for all four evaluation metrics. One exception was the ODIR-5K dataset, where the LMF-loss had the second-best performance, with the BWCCE loss performing just slightly better than the LMF-loss.
\end{flushleft}
\label{tab:table4}
\end{table}

For the ODIR-5K test set, we can see that our proposed method achieved 62.94, 57.73, and 59.97 macro-f1 scores on EfficentNetV2, ResNet50, and DenseNet121, respectively. The proposed LMF loss showed up to 2.5\%-3\% performance improvement when compared to the CCE, Focal, and LDAM loss functions. However, the recently proposed BWCCE loss slightly outperformed the proposed method on EffecientNetV2 and DenseNet121 models by achieving macro-f1 scores of 63.41 and 60.39, respectively. Even though BWCCE loss outperformed LMF loss by a small margin in some cases for the ODIR-5K dataset, LMF loss still showcased significant improvements over the other three methods.

The proposed LMF loss also demonstrated significant improvements for both skin-cancer datasets used in this study. For the HAM-10K test set, the LMF loss framework showed a performance improvement of around 2\%-4\% when compared to all other methods. It achieved macro-f1 scores of 84.61, 82.43, and 83.31 on the EfficentNetV2, ResNet50, and DenseNet121 models, respectively. We can see the most significant improvement in the ISIC-2019 dataset. Our proposed framework achieved an improvement of up to almost 9\% when compared to some of the other baselines. The CCE and Focal loss performed poorly on the larger EfficientNetV2 and ResNet50 architectures for the ISIC dataset. However, the LMF loss showed around 5\%-9\% performance improvement compared to these loss functions on both architectures. The proposed method also achieved about 2-6\% performance improvement when compared to the LDAM and BWCCE loss on all three architectures. The LMF loss achieved the best macro-f1 scores of 80.15, 77.68, and 76.27 on EfficentNetV2, ResNet50, and DenseNet121, respectively.

On the other hand, all loss functions performed significantly well on the Covid-19 chest X-ray dataset by achieving significantly higher macro-f1 scores when compared to the other datasets. In addition, for the Covid-19 chest X-ray test set, the LMF loss achieved the highest macro-f1 scores of 97.38, 96.50, and 97.06 on EfficentNetV2, ResNet50, and Dense-Net121, respectively.

It also is worth noting that in the other metrics, such as accuracy, precision, and recall, the LMF loss outperformed others in most cases. From \textbf{Tables \ref{tab:table2}, \ref{tab:table3}, and \ref{tab:table4}}, we can also observe that the focal loss performed poorly on the ISIC-2019 dataset and occasionally scored lower on the f1 score than the standard CCE loss. In contrast, the LDAM loss showcased some notable improvements compared to the standard CCE loss. However, by applying our proposed LMF loss framework, which consists of both the focal and LDAM loss, the models achieved significant performance improvement, beating the f1 scores obtained from those loss functions individually. This improvement demonstrated how utilizing both loss functions can be beneficial for enhancing model performance.

\textbf{Fig \ref{fig1}} gives an overview of how the five methods performed across all three architectures used in this study. For the ODIR-5K dataset, the proposed LMF loss scored a higher mean macro-f1 score than the BWCCE loss, even though the BWCCE loss scored slightly higher individual f1 scores on the EfficientNetV2 and DenseNet-121 models. In contrast to the LMF loss, the BWCCE loss showed a much higher deviation in the macro-f1 score across multiple models. For the HAM-10K and the ISIC-2019 dataset, the LMF loss achieved about a 2-7\% improvement in the mean macro-f1 score when compared to other loss functions. The LMF loss also achieved a marginally higher mean macro-f1 score for the COVID-19 dataset when compared to other loss functions. Overall, we can see from \textbf{Fig \ref{fig1}} that our proposed framework achieved a higher average macro-f1 score and a moderate deviation in macro-f1 scores across multiple architectures and datasets. This demonstrates the performance consistency of the proposed framework across multiple architectures.

In addition, we further investigated the performance of our proposed method at the class level to demonstrate the performance of our proposed framework on minority classes. Here, we present the findings from the EfficientNetV2 model for additional investigation because it outperformed the other models in terms of performance. \textbf{Table \ref{tab:table5}} showcases per class f1 scores of all four test sets for the EfficientNetV2 architecture. 

% % Place tables after the first paragraph in which they are cited.
\begin{table}[!ht]
\begin{adjustwidth}{-2.25in}{0in} % Comment out/remove adjustwidth environment if table fits in text column.
\centering
\caption{
{\bf Class-wise analysis of f1 scores on all four test sets for EfficientNetV2.}}
\begin{NiceTabular}{l|l|c|c|c|c|c|c}
\toprule
\textbf{Dataset} & \textbf{Class} & \textbf{Train Samples} & \textbf{CCE} & \textbf{Focal} & \textbf{LDAM} & \textbf{BWCCE}  & \textbf{LMF (ours)}\\

\midrule
ODIR-5K & A & 186 & \textbf{70.13} & 60.53 & 59.74 & 66.67 & 57.97\\
\cmidrule{2-8}
& C & 205 & 83.72 & 82.11 & 80.41 & \textbf{85.71} & 84.21\\
\cmidrule{2-8}
& D & 1126 & 64.10 & 62.21 & \textbf{65.49} & 62.58 & 65.12\\
\cmidrule{2-8}
& G & 198 & 52.78 & \textbf{64.20} & 46.34 & 61.97 & 55.70\\
\cmidrule{2-8}
& H & 90 & 8.33 & 13.33 & 13.79 & 24.24 & \textbf{28.57}\\
\cmidrule{2-8}
& M & 162 & 85.33 & 87.67 & 86.49 & 82.67 & \textbf{91.89}\\
\cmidrule{2-8}
& N & 2011 & 73.67 & 73.74 & 72.65 & 74.49 & \textbf{76.75}\\
\cmidrule{2-8}
& O & 496 & 42.58 & 42.94 & 34.52 & \textbf{48.91} & 43.27\\

\midrule
HAM-10K & AKIEC & 229 & 72.34 & 72.00 & 70.45 & 75.56 & \textbf{79.57}\\
\cmidrule{2-8}
& BCC & 360 & 84.97 & 82.28 & 83.66 & 86.09 & \textbf{90.68}\\
\cmidrule{2-8}
& BKL & 769 & 78.64 & 78.64 & 77.32 & \textbf{80.62} & 79.64\\
\cmidrule{2-8}
& DF & 81 & 71.43 & 75.86 & \textbf{81.25} & 77.41 & 80.00\\
\cmidrule{2-8}
& MEL & 779 & 74.05 & 71.21 & 71.21 & 71.43 & \textbf{77.99}\\
\cmidrule{2-8}
& NV & 4693 & 95.02 & 94.33 & 95.47 & 95.30 & \textbf{96.00}\\
\cmidrule{2-8}
& VASC & 100 & 95.24 & 90.91 & \textbf{95.45} & 87.50 & 88.37\\
\midrule

ISIC-2019 & AK & 607 & 63.36 & 58.06 & \textbf{68.85} & 65.08 & 68.12\\ \cmidrule{2-8}
& BCC & 2327 & 85.19 & 85.11 & 88.18 & 86.00 & \textbf{88.58}\\
\cmidrule{2-8}
& BKL & 1836 & 75.53 & 71.27 & 77.50 & 75.50 & \textbf{77.68}\\
\cmidrule{2-8}
& DF & 167 & 66.67 & 55.88 & 73.85 & 74.19 & \textbf{78.26}\\
\cmidrule{2-8}
& MEL & 3166 & 74.28 & 70.06 & 75.46 & 75.47 & \textbf{77.30}\\
\cmidrule{2-8}
& NV & 9013 & 90.84 & 88.27 & 91.68 & 90.14 & \textbf{92.09}\\
\cmidrule{2-8}
& SCC & 440 & 62.70 & 59.18 & 65.19 & 63.10 & \textbf{69.27}\\
\cmidrule{2-8}
& VASC & 177 & 88.31 & 86.08 & 87.50 & 89.47 & \textbf{90.00}\\

\midrule
Covid-19 & Covid & 2,532 & 99.08 & 97.69 & 99.35 & 99.08 & \textbf{99.53}\\ \cmidrule{2-8}
& LO & 4,208 & 92.88 & 92.70 & 93.85 & 93.15 & \textbf{94.26}\\ 
\cmidrule{2-8}
& Normal & 7,134 & 96.02 & 95.79 & 96.47 & 96.11 & \textbf{96.72}\\
\cmidrule{2-8}
& VP & 941 & \textbf{99.26} & 97.79 & 98.77 & 99.01 & 99.01\\
\bottomrule

\end{NiceTabular}
\begin{flushleft} The proposed LMF loss showcased consistent performance improvements for almost all individual classes when compared to other baselines. In particular, the proposed method achieved the best f1 scores for 7 out of the 8 classes in the ISIC-2019 dataset and 3 out of the 4 classes in the Covid-19 dataset.
\end{flushleft}
\label{tab:table5}
\end{adjustwidth}
\end{table}

From \textbf{Table \ref{tab:table5}}, we can see that the proposed LMF loss and the BWCCE loss demonstrated good class-wise f1 scores on the ODIR-5K dataset. We can also see from the table that classes H and M had the least number of training samples in the ODIR-5K dataset. With only 90 samples in class H, the LMF loss achieved an f1 score of 28.57, which was significantly better than the other methods. Also, in class M, where the total number of training samples was 162, our LMF loss framework achieved a 91.89 f1 score with more than 4\% improvement over the second-highest score obtained by focal loss. Additionally, LMF loss obtained the highest f1 score of 76.75 for class N. On the other hand, the BWCCE loss achieved the best f1 scores of 85.71 and 48,91 for classes C and O, respectively.

In the HAM dataset, the DF class was the minority class with only 81 training samples. We can see from \textbf{Table \ref{tab:table5}}, the LDAM loss achieved the best f1 score of 81.25 for the DF class. However, the LDAM loss failed to showcase similar performance for other classes. In comparison, the proposed LMF loss framework demonstrated consistent performance improvements across all classes. It achieved the second-best f1 score of 80\% in the smallest DF class. In addition, LMF loss also achieved the best f1 scores of 79.57 and 90.68 for the minority classes: AKIEC and BCC, which was a little over 4\% improvement over the second-highest f1 scores achieved by the BWCCE loss. Overall, the proposed LMF loss achieved the best f1 scores for 4 of the 7 classes in the HAM dataset.

The least number of training samples in the ISIC-2019 dataset was in class DF, with 167 samples. The LMF loss framework achieved a 78.26 f1 score, which was almost a 12\% improvement over the standard CCE loss. Whereas the second-highest score of 74.19 was obtained from the BWCCE loss, which was still 4\% less than the LMF loss. LMF loss also outperformed other methods by achieving 90.00 and 69.27 f1 scores for the minority classes VASC and SCC, respectively. Overall, the proposed method achieved the best f1 scores for 7 of the 8 classes present in the ISIC dataset. We also saw similar improvements for the COVID-19 Radiography Dataset. The LMF loss achieved the best f1 scores for 3 of the 4 classes present in the Covid-19 dataset. These results further demonstrate the consistency of the proposed framework.

\subsection*{Attention map analysis}
Here, we present our qualitative results using Grad-CAM \cite{Selvaraju2016} attention map comparison. We annotated our sample images with the help of a medical practitioner. We instructed the annotator to identify the important regions of interest within the images using small bounding boxes. To eliminate bias, the annotator did not view the attention maps before making the annotations. It is worth noting that in some cases, particularly with images from skin cancer datasets, the annotator stated that it is challenging for a doctor to make a diagnosis based solely on an image of a small patch of skin. Typically, a doctor would examine other parts of the patient's body and consider symptoms or lab tests for a diagnosis. Therefore, in these instances, the annotator provided annotations based on their best judgment. Nevertheless, these annotations helped us determine whether the models concentrated on important characteristics within the pictures.

\begin{figure}[!h]
\begin{adjustwidth}{-2.25in}{0in} 
    \includegraphics[width=\linewidth]{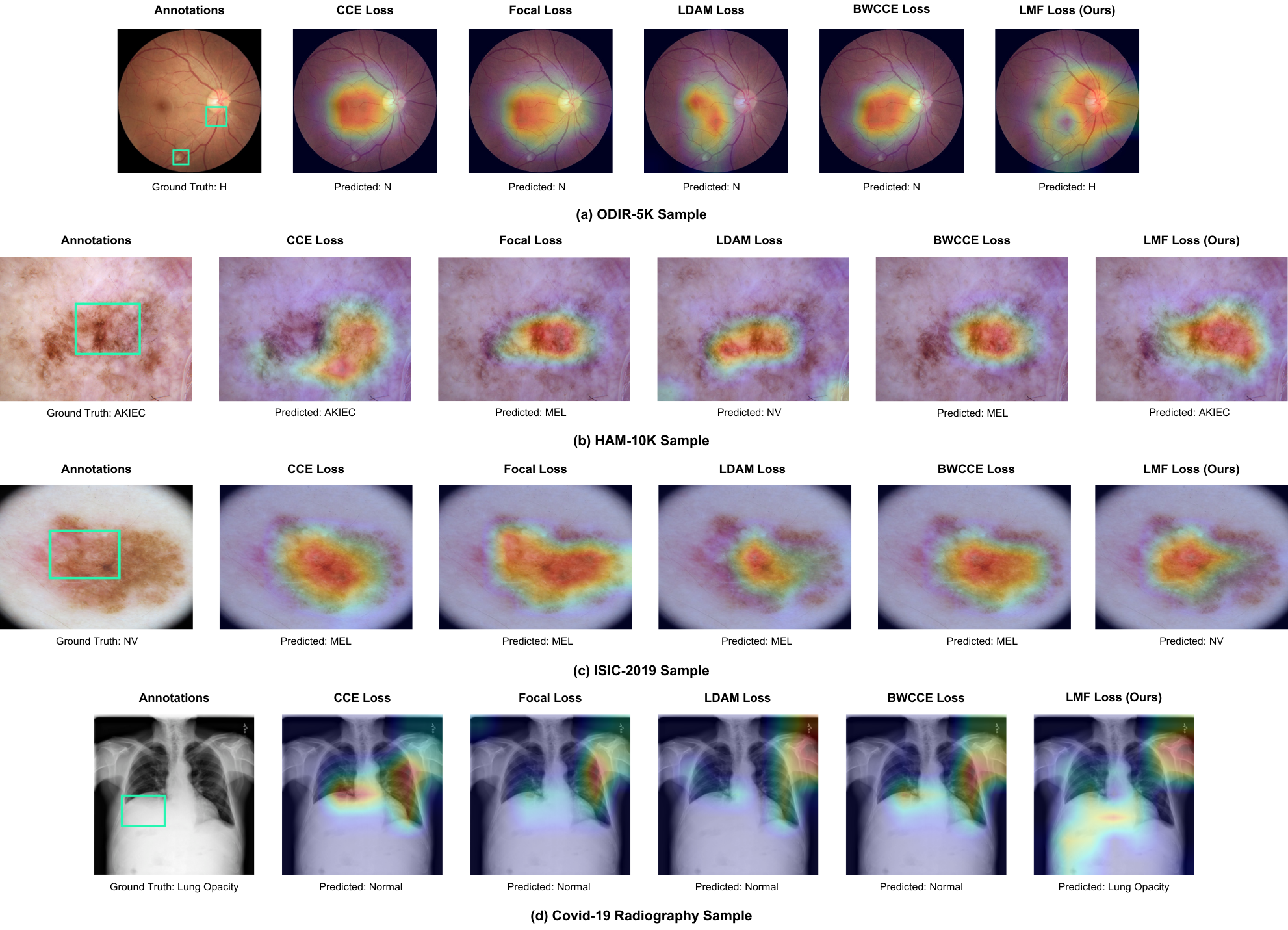}
    \caption{{\bf Grad-CAM attention map visualization and comparison on test samples from all four datasets.} Green bounding boxes depict annotations obtained from a doctor.}
    \label{fig3}
\end{adjustwidth}
\end{figure}

\textbf{Fig \ref{fig3}A} shows attention maps generated for a sample from minority class H (Hypertension) of the ODIR-5K dataset. We can see that the CCE, Focal, LDAM, and BWCCE loss functions misclassified the image as N (Normal) class. In all four cases, we can see that the generated heatmap barely concentrated on the annotated locations. In comparison, the heatmap from the model trained with LMF loss had much better coverage on the annotated regions and predicted the accurate class. In \textbf{Fig \ref{fig3}B}, we can see that the attention maps generated from Focal, LDAM, and BWCCE loss did have some decent coverage on the annotated region. However, all three methods provided incorrect predictions for the disease class. The CCE loss accurately predicted the class, but it only focused on a portion of the features from the annotated region. In comparison, the LMF concentrated on most of the features within the annotated region and correctly predicted the class as Actinic keratoses (AKIEC).

For the ISIC-2019 sample in \textbf{Fig \ref{fig3}C}, we can see the LMF loss trained model was able to accurately predict the class as NV (Melanocytic nevus). We can see that LMF loss generated a very concise attention map that perfectly focused on the characteristics within the annotated bounding box. Whereas the CCE, Focal, and BWCCE loss generated large attention areas focusing on too many features, possibly causing them to misclassify the image as MEL (Melanoma). Lastly, \textbf{Fig \ref{fig3}D} shows that the attention maps from the baseline loss functions barely concentrated on the primary region of interest within the original image. Thus resulting in them labeling the image as normal. In contrast, the attention map from LMF loss had good coverage around the annotated region and accurately predicted the class as lung opacity (LO).

\subsection*{Analysis of LMF loss hyperparameters}
As we mentioned previously, the proposed LMF loss contains two hyperparameters; \textalpha~ and \textbeta. Through our experiments, we found that keeping the hyperparameter values between 0.5 to 2.0 yielded the best results. We explored the impact of the hyperparameters further by analyzing the effect of \textalpha~ and \textbeta~ on the HAM-10K dataset. To perform the analysis, we split the hyperparameter settings into three categories-

\begin{enumerate}
    \item \textbf{Setting-1:} We modified the \textalpha~ and \textbeta~ parameters simultaneously, and they were equal to each other. For instance, we trained a model with \textalpha=0.5, \textbeta=0.5, then another model with \textalpha=0.7, \textbeta=0.7, and so on. We trained a total of eleven models in this category, with the \textalpha, \textbeta~values of LMF-loss ranging from 0.5 to 2.0.

    \item \textbf{Setting-2:} We only modified the \textbeta~ parameter of the loss function while \textalpha=1.0. For instance, we trained a model \textalpha=1.0, \textbeta=0.5, then another model with \textalpha=1.0, \textbeta=0.7, and so on. We also trained eleven models with these hyperparameter settings, with the \textbeta~ value of  LMF-loss ranging from 0.5 to 2.0.

    \item \textbf{Setting-3:} We only modified the \textalpha~ parameter of the loss function while \textbeta=1.0. For instance, we trained a model with \textalpha=0.5, \textbeta=1.0, then another model with \textalpha=0.7, \textbeta=1.0, and so on. We trained a total of eleven models with these hyperparameter settings, with the \textalpha~ value of  LMF-loss ranging from 0.5 to 2.0.
\end{enumerate}

\begin{figure*}[!h]
    \includegraphics[width=\linewidth]{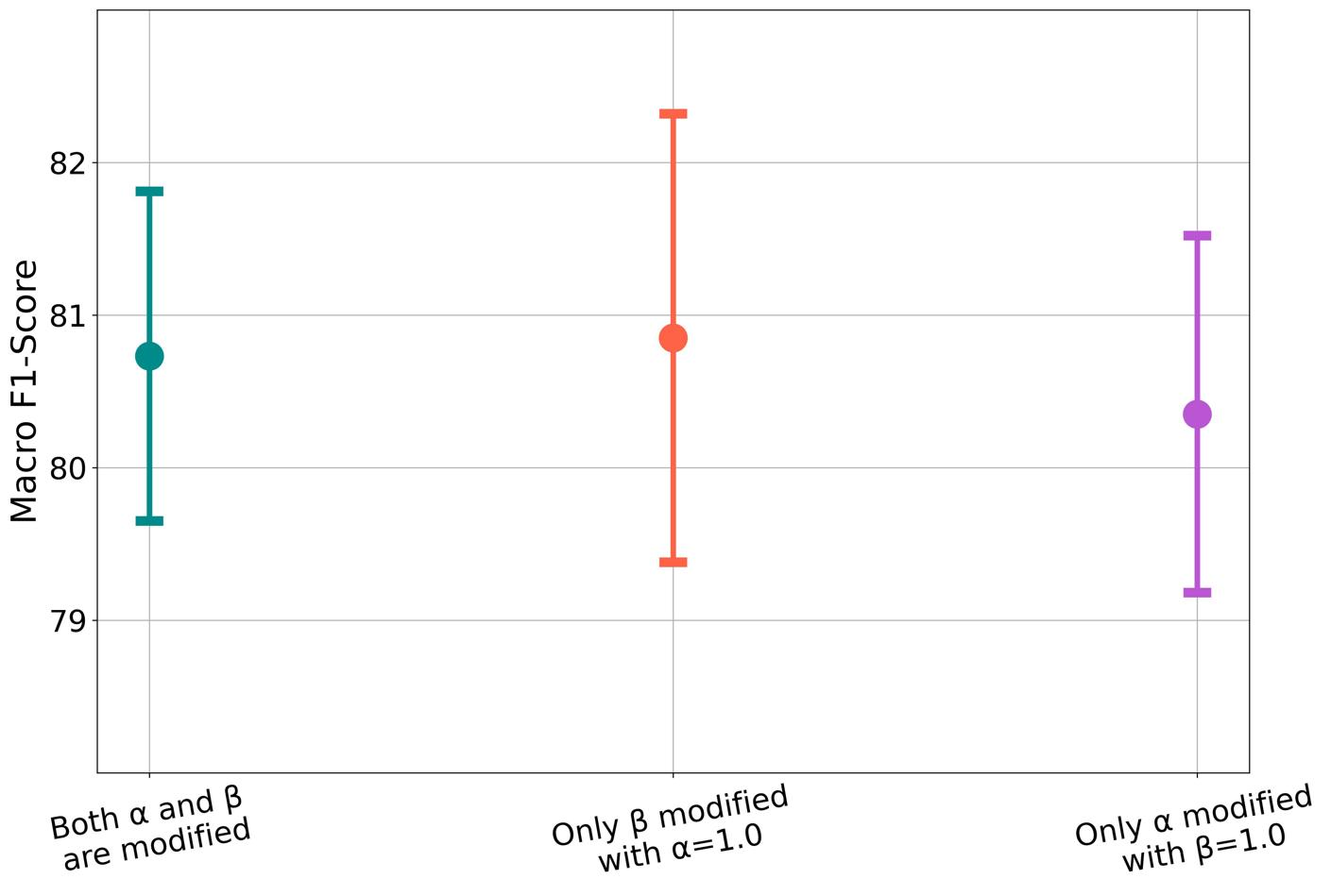}
    \caption{{\bf Mean macro-f1 scores obtained from each hyperparameter setting, along with their average deviation.} We obtained the highest average macro-f1 score when we only modified the \textbeta~ value of the LMF-loss, and \textalpha~ was set to 1.0.}
    \label{fig4}
\end{figure*}

All experiments were performed on the EfficientNetV2 model using the aforementioned hyperparameter settings. Due to the large number of training runs required for the analysis, we only trained each model for 60 epochs. \textbf{Fig \ref{fig4}}, presents the mean macro-f1 scores obtained from each hyperparameter setting and their average deviation. \textbf{Fig \ref{fig4}} shows that the highest mean macro-f1 score was achieved in setting-2 when we only modified the \textbeta~ value, and \textalpha~ was equal to 1.0. However, setting-1 showcased the least deviation in the macro-f1 score with its tighter error margins. Setting-3, where we only modified the \textalpha~ value, showcased the least mean macro-f1 score with moderate deviations.

We can also see the patterns from our hyperparameter analysis in the results presented for LMF-loss in the previous section. \textbf{Table \ref{tab:table6}} showcases the \textalpha~ and \textbeta~ values we used to obtain the best results for the proposed framework. In \textbf{Table \ref{tab:table6}}, we can see that we achieved most of the best results using the hyperparameter setting-2, where only the \textbeta~ value was modified while \textalpha~=1.0. We found that choosing a \textbeta~ value of 0.5 or 2.0 was a good starting point, as 6 out of the 12 best results we obtained were using these two \textbeta~values. Overall, we found using the hyperparameter setting-2 to be most beneficial for the datasets used in this study. However, these hyperparameter settings may vary with different datasets.

\begin{table}[!ht]
\centering
\caption{{\bf \textalpha~ and \textbeta~ values that generated the best results for LMF loss.}}

\begin{NiceTabular}{l| c | c | c | c}
\toprule
\textbf{Dataset} & \textbf{\makecell{Hyper-\\parameter}} & \textbf{EfficientNetV2} & \textbf{ResNet-50} & \textbf{DenseNet-121}\\ 
\midrule

ODIR-5K & \textalpha & 1.0 & 1.0 & 1.0\\ \cmidrule{2-5}
& \textbeta & 0.6 & 0.5 & 1.1\\
\midrule

HAM-10K & \textalpha  & 1.0 & 0.5 & 1.0\\ \cmidrule{2-5}
& \textbeta & 2.0 & 1.0 & 1.3\\
\midrule

ISIC-2019 & \textalpha  & 1.0 & 1.0 & 1.0\\ \cmidrule{2-5}
& \textbeta & 0.5 & 0.5 & 2.0\\
\midrule

Covid-19 & \textalpha & 1.0 & 1.0 & 1.0\\ \cmidrule{2-5}
& \textbeta & 1.5 & 1.2 & 2.0\\
\bottomrule
\end{NiceTabular}

\label{tab:table6}
\end{table}

\section*{Conclusion}

In order to address the imbalance issue in medical image classification, we present a novel method named Large Margin aware Focal (LMF) loss, which integrates the focal loss and the LDAM loss into a single hybrid framework. The proposed method employs a linear combination of these two loss functions, weighted by two hyperparameters. The framework combines the strengths of both loss functions by imposing larger margins for the minority classes and dynamically emphasizing the difficult samples present in the datasets. Through a comprehensive evaluation of medical image classification datasets from diverse domains, including ocular disease diagnosis (ODIR-5K), skin cancer diagnosis (HAM-10K and ISIC-2019), and covid-19 diagnosis (Covid-19 Radiography), we compared the proposed framework to baseline loss functions such as CCE loss, Focal loss, LDAM loss, and BWCCE loss. Our experiments encompassed three popular neural network architectures: EfficientNetV2, ResNet-50, and DenseNet-121. The results consistently demonstrated the superior performance of the proposed framework across all datasets and architectures, setting it apart from the other loss functions, which struggled to perform consistently across various datasets and architectures. Notably, the LMF loss framework demonstrated a noteworthy enhancement in macro-f1 scores, ranging from 2\% to 9\%, across a diverse range of test cases. These consistent improvements were observed across multiple evaluation metrics and were further supported by detailed attention map comparisons. We hope future researchers will greatly benefit from utilizing our simple yet reliable framework. We also envision extending the application of our proposed method to other imbalanced medical imaging problems, such as image segmentation.

\section*{Acknowledgments}
We would like to thank Doctor Lubaba Binte Saber for providing us with the annotations required for attention map analysis. She obtained her Bachelor of Medicine and Bachelor of Surgery (MBBS) degree from Sir Salimullah Medical College, Dhaka, Bangladesh.

\section*{Data availability statement}
The source code of our implementations are publicly available at: \url{https://github.com/Adnan-Sadi/LMFLOSS}.

All datasets used in this study are also publicly available in the following online repositories:
\begin{enumerate}
    \item ODIR-5K: \url{https://www.kaggle.com/datasets/andrewmvd/ocular-disease-recognition-odir5k}
    \item HAM-10K: \url{https://www.kaggle.com/datasets/surajghuwalewala/ham1000-segmentation-and-classification}
    \item ISIC-2019: \url{https://www.kaggle.com/datasets/andrewmvd/isic-2019}
    \item COVID-19 Radiography: \url{https://www.kaggle.com/datasets/tawsifurrahman/covid19-radiography-database}
\end{enumerate}

\nolinenumbers

\bibliography{bibliography}

\end{document}